\documentclass[10pt,twocolumn,letterpaper]{article}

\usepackage[dvipsnames,table,xcdraw]{xcolor}

\usepackage{cvpr}              

\usepackage[numbers,sort&compress]{natbib}

\usepackage{graphicx}
\usepackage{amsmath}
\usepackage{amssymb}
\usepackage{booktabs}

\usepackage{float}
\usepackage{xspace}
\usepackage{epsfig}

\usepackage{amsmath,amsfonts,bm}

\def\eqref#1{equation~\ref{#1}}

\def\1{\bm{1}}

\def\rvs{{\mathbf{s}}}

\def\rvx{{\mathbf{x}}}

\def\rmZ{{\mathbf{Z}}}

\def\ermZ{{\textnormal{Z}}}

\DeclareMathAlphabet{\mathsfit}{\encodingdefault}{\sfdefault}{m}{sl}
\SetMathAlphabet{\mathsfit}{bold}{\encodingdefault}{\sfdefault}{bx}{n}

\newcommand{\E}{\mathbb{E}}

\usepackage{nccmath}

\usepackage{multirow,booktabs}
\usepackage{colortbl}
\usepackage{tabularx}
\usepackage{url}
\usepackage{algorithm,algpseudocode}

\usepackage{enumitem}
\setlist{topsep=0pt, leftmargin=*}

\algdef{SE}[SUBALG]{Indent}{EndIndent}{}{\algorithmicend\algorithmicend\ \quad}%
\algtext*{Indent}
\algtext*{EndIndent}

\newlength\myindent 
\usepackage{xpatch}

\makeatletter
\xpatchcmd{\algorithmic}
  {\ALG@tlm\z@}{\leftmargin\z@\ALG@tlm\z@}
  {}{}
\makeatother

\algrenewcommand\algorithmicrequire{}

\newcommand{\method}{NMD\xspace}
\newcommand{\methodlong}{neural mean discrepancy\xspace}

\newcommand{\fakeparagraph}[1]{\vspace{1mm}\noindent\textbf{#1}}
\usepackage[pagebackref,breaklinks,colorlinks]{hyperref}

\usepackage[capitalize]{cleveref}
\crefname{section}{Sec.}{Secs.}
\Crefname{section}{Section}{Sections}
\Crefname{table}{Table}{Tables}
\crefname{table}{Tab.}{Tabs.}

\definecolor{myblue}{RGB}{150, 168, 207}
\definecolor{mypink}{RGB}{231, 138, 195}
\definecolor{myred}{RGB}{255,33,33}
\definecolor{mygreen}{RGB}{33,144,33}

\def\cvprPaperID{186} 
\def\confName{CVPR}
\def\confYear{2022}

\begin{document}

\title{Neural Mean Discrepancy for Efficient Out-of-Distribution Detection}

\author{Xin Dong$^{1}$, 
Junfeng Guo$^{2}$, 
Ang Li$^{23}$,
Wei-Te Ting$^{1}$, 
Cong Liu$^{2}$,
H.T. Kung$^{1}$\\
\vspace{-0.35cm}
\\
$^1$Harvard University,\quad$^2$UT Dallas,\quad$^3$Google DeepMind\\
\tt\small xindong@g.harvard.edu
}
\maketitle

\begin{abstract}

Various approaches have been proposed for out-of-distribution~(OOD) detection by augmenting models, input examples, training sets, and optimization objectives. Deviating from existing work, we have a simple hypothesis that standard off-the-shelf models may already contain sufficient information about the training set distribution which can be leveraged for reliable OOD detection. Our empirical study on validating this hypothesis, which measures the model activation's mean for OOD and in-distribution~(ID) mini-batches, surprisingly finds that activation means of OOD mini-batches consistently deviate more from those of the training data. In addition, training data's activation means can be computed offline efficiently or retrieved from batch normalization layers as a `free lunch'. Based upon this observation, we propose a novel metric called Neural Mean Discrepancy (NMD), which compares neural means of the input examples and training data. Leveraging the simplicity of NMD, we 
propose an efficient OOD detector that computes neural means by a standard forward pass followed by a lightweight classifier. Extensive experiments show that \method outperforms state-of-the-art OOD approaches across multiple datasets and model architectures in terms of both detection accuracy and computational cost.      
\end{abstract}
 
\section{Introduction}
\label{sec:intro}

Deep Neural Networks (DNNs) have achieved successes on many computer vision tasks~\cite{lecun2015deep,goodfellow2016deep}. However, most of the deep learning methods are based on an assumption that the data is independent and identically distributed (\emph{i.i.d.}), \ie, training and testing data come from the same underlying distributions. While it is almost impossible to curate a dataset that covers all different kinds of scenarios in the real world, the \emph{i.i.d.} assumption is untrue in practice and out-of-distribution (OOD) examples are likely to occur in the testing data. 
So the ability to detect OOD examples becomes essential when deploying deep neural networks in real-world applications~\cite{2110.11334,salehi2021unified}.

\begin{figure}[t]
    \centering
    \includegraphics[width=0.75\linewidth]{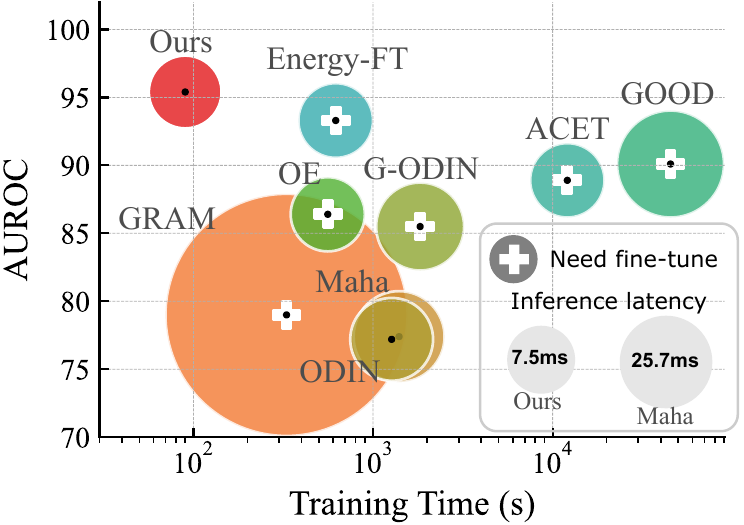}
    \caption{Training and inference time comparison with CIFAR-10 against CIFAR-100 (OOD) detection on ResNet-34. Our~\emph{NMD-MLP} achieves superior performance in terms of both AUROC and training time. See~\cref{sec:method_train_ood,sec:training_inference_exp,fig:compare_in_out_data} for more details.}
    \label{fig:latency}
\end{figure}

Many approaches have been developed to address OOD examples including enhancing standard DNN architectures~\cite{gal2016dropout,lakshminarayanan2017simple,vyas2018out,chen2020oodanalyzer,devries2018learning,havasi2020training,ren2019likelihood} and DNN fine-tuning using the augmented training set~\cite{lee2017training,chen2020informative,mohseni2020self,liu2020energy,karandikar2021soft}. Unfortunately, these methods often incur significant overhead \emph{w.r.t.} both computation and data processing. Recent studies perform kernel density estimation on standard training sets, interpreting the negative of incoming example's density as the outlier score~\cite{guillory2021predicting,erdil2021taskagnostic,morningstarDensityStatesEstimation2021,jia2021efficient}. Both non-parametric and parametric kernels have been studied in the literature. However, they suffer from limited performance, heavy reliance on large batch size, and low computational efficiency.

Deviating from most previous works, we believe the off-the-shelf model itself should contain sufficient information about the training data distribution. So we proposed a simple study (\Cref{fig:scatter}) by looking at the model activation's mean for OOD and ID input batches. The result reveals that the activation means of OOD mini-batches consistently and clearly deviate more from those of the training data. Inspired by this observation, we raised the question: \emph{Can OOD detection be as simple and efficient as computing activation's arithmetic mean without fine-tuning?}

We propose a novel metric called \emph{Neural Mean Discrepancy (NMD)}, which compares neural means of the input examples and training data. The proposed NMD metric can be efficiently computed from the model's activations; only forward passes are needed. Additionally, training data's neural mean can be obtained for free from Batch Normalization layers~\cite{ioffe2015batch}.  We found this NMD metric able to achieve superior performance in OOD detection in terms of both accuracy and efficiency (\Cref{fig:latency}).

From a theoretical perspective, we further connect the aforementioned observation and the NMD formulation with integral probability metrics~(IPMs).
IPMs are  a family of general distribution distance metrics, which project two sets of examples to a new space via a kernel and use the mean discrepancy of their projections as the distribution distance. 
Both non-parametric and deep neural kernels have been studied in the past~\cite{guillory2021predicting,morningstarDensityStatesEstimation2021,jia2021efficient}. The key finding of our work is that, instead of defining a separate kernel function, the off-the-shelf DNN itself is an efficient and effective kernel for the purpose of out-of-distribution detection. This finding consequently brings several advantages of our approach summarized as follows:
\vspace{1mm}
\begin{enumerate}
\itemsep0em 
    \item \emph{Accessibility:} Since the off-the-shelf DNN can be directly used, our NMD distance metric does not require data- and computation-intensive kernel optimization, fine-tuning, or hyper-parameter search.
    \item \emph{Extensibility:} Each group of neurons (\eg, each channel in a convolution layer) are treated as a unique kernel, which allows for thousands of parallelized kernels. They are from different depths of the DNN and  complementary to each other for capturing multi-level semantics, which leads to  improved discriminatory power.
    \item \emph{Simplicity.} Computing the NMD metric turns out, surprisingly, as simple as calculating DNN's activation means. It can be offline computed via forward passes on the training data. Interestingly, if the model contains Batch Normalization (BN) layers, the neural means can be approximated from BN directly as a ``\textit{free lunch}''.
\end{enumerate}
\vspace{1mm}

\begin{figure}[!t]
    \centering
    \includegraphics[width=0.8\linewidth]{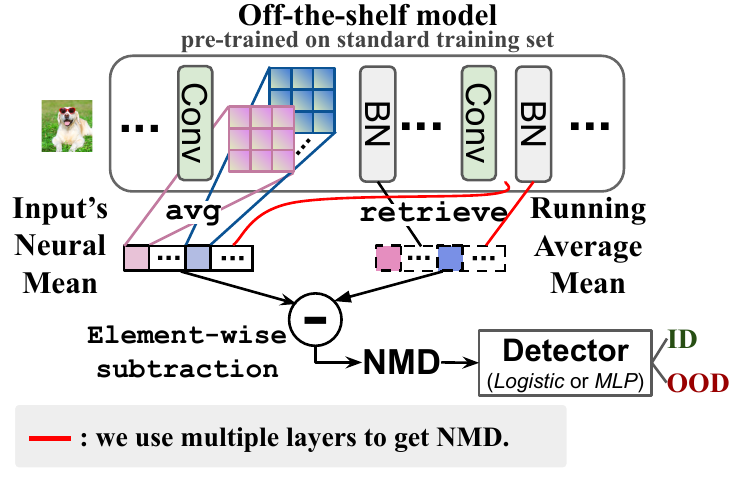}
    \caption{The pipeline of \method-based OOD detection. An input example's \method vector is computed by taking the difference between its channel-wise activation mean and corresponding running average in the batch normalization~(BN) layer. The \method vector is then passed to a lightweight classifier (\eg, LR or MLP). Please be advised that BN is not a requirement in computing \method.}
    \label{fig:pipeline}
\end{figure}

\begin{figure}[t]
    \centering
    \includegraphics[width=0.9\linewidth]{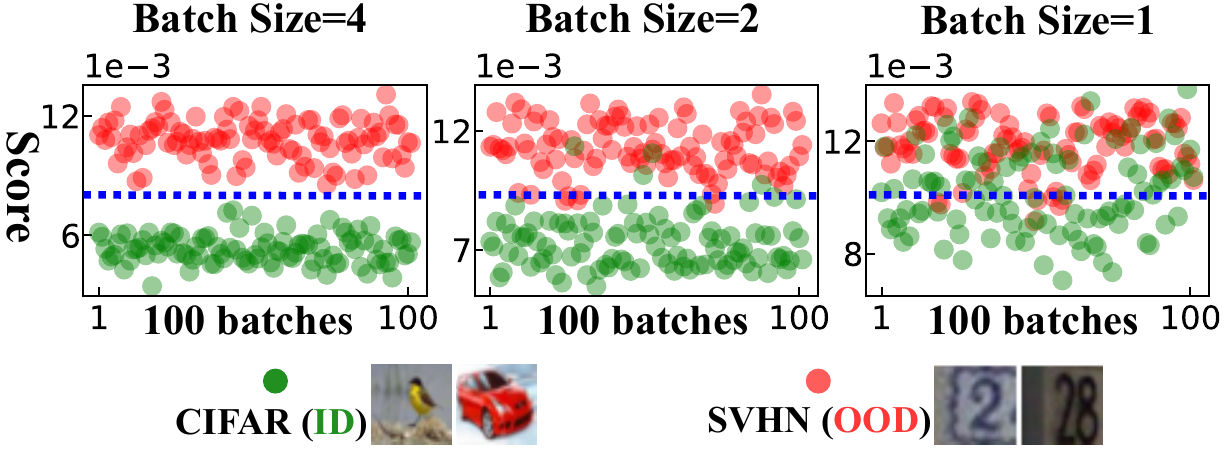}
    \caption{A proof-of-concept example for an off-the-shelf ResNet-34 pretrained on CIFAR-10 (ID). We first compute each mini-batch's \method (see \Cref{fig:pipeline}). We then take the average of elements' magnitude in \method vector for each mini-batch as the score (y-axis), referred to as \textit{Ours-Avg}. Without fine-tuning, \textit{Ours-Avg} can reliably separate ID and OOD data. However, an un-trained ResNet-34 is not able to achieve this as shown in the appendix.}
    \label{fig:scatter}
\end{figure}

We find the absolute value of \method is able to reliably distinguish ID against OOD batches even when the batch size is down to 4, an order of magnitude smaller than previous statistical methods~\cite{grosse2017statistical,carlini2017adversarial,gao2020maximum,guillory2021predicting,jia2021efficient}. In order to further improve the detection efficacy, we introduce a lightweight OOD detector~(instantiated as either a logistic regression or a multilayer perceptron) which takes neural means as the input to generate detection outputs. The detector is able to take sensitivity and correlation of elements in the \method vector into consideration, and achieve state-of-the-art detection accuracy even when the batch size becomes 1, \ie, single example OOD detection. The entire pipeline of our method is illustrated in \Cref{fig:pipeline} and \cref{alg:pipeline}.

We extensively evaluate \method across various datasets, types of OOD (\textit{far}- and \textit{near}- OOD), pre-training types~(supervised and self-supervised~\cite{he2020momentum}), and model architectures~(Simple ConvNet~\cite{jia2021efficient}, ResNet~\cite{he2016deep}, VGG~\cite{simonyan2014very} and Vision Transformer~\cite{dosovitskiy2020vit}). \method consistently outperforms statistical approaches and other state-of-the-art methods in these settings. 
We further evaluate the robustness and generalizability of \method under various data circumstances including few-shot ID and OOD examples, zero-shot OOD examples, and transfer learning for unseen OOD. 
In addition, we measure the efficiency of our approach showing that the training cost of an \method detector is orders of magnitude faster than existing methods~\cite{lee2018simple,2110.11334,liu2020energy,bitterwolf2020certifiably}, and our overall inference latency is close to a standard forward pass.


\section{Preliminary}
\subsection{Out-of-distribution (OOD) detection}
Suppose one has a model well-trained on the training set $\mathcal{D}_\texttt{tr}=\{\rvs_1, \dots, \rvs_{|\mathcal{D}_\texttt{tr}|}\}$ from an underlying distribution $\mathbb{P}$.
Given a batch of input examples $\mathcal{I}=\{\rvx_1, \dots, \rvx_{|\mathcal{I}|}\}$ from an unknown distribution $\mathbb{Q}$, the goal of OOD detection is to discriminate whether $\mathcal{I}$ comes from $\mathbb{P}$ in a similar spirit to measure how far $\mathbb{Q}$ deviates from $\mathbb{P}$. 

\subsection{Integral probability metrics}
Integral probability metrics (IPMs)~\cite{muller1997integral} is a family of probability distance measures defined as
\begin{equation}
    \texttt{IPM}_{\mathcal{F}}(\mathbb{Q},\mathbb{P})=\sup_{\varphi\in \mathcal{F}}\  \left({\E}_{\rvx\sim \mathbb{Q}}[\varphi(\rvx)]-{\E}_{\rvs\sim \mathbb{P}}[\varphi(\rvs)]\right), 
    \label{equ:ipm_theory}
\end{equation}
where $\varphi(\cdot)$ denotes the witness function. IPMs project the examples from two distributions $\mathbb{P}, \mathbb{Q}$ to a new space using $\varphi$, and then compare the means of the two projected sets. 
Normally, we do not know the exact distribution formulation, thus \cref{equ:ipm_theory} is empirically estimated as
\begin{equation}
    \sup_{\varphi\in \mathcal{F}}\ \left(\frac{1}{|\mathcal{I}|}\sum_{i=1}^{|\mathcal{I}|}\varphi(\rvx_i)-\frac{1}{|\mathcal{D}_\texttt{tr}|}\sum_{j=1}^{|\mathcal{D}_\texttt{tr}|}\varphi(\rvs_j)\right).
    \label{equ:ipm_emp}
\end{equation}
If $\mathcal{I}$ is an out-of-distribution batch, we expect the value of \cref{equ:ipm_emp} to be large; otherwise, it should be relatively small.

IPM is a general framework, which relies on choosing an appropriate class of witness functions $\mathcal{F}$. 
Although IPM-based methods have theoretical guarantees, they have certain limitations: (1) They may be incapable of handling high
dimensional data like images~\cite{kirchler2020two} or capture semantic information~\cite{carlini2017adversarial,liu2020learning}. (2) They usually rely on hypothesis testing which requires sufficiently large $|\mathcal{I}|, |\mathcal{\mathcal{D}_\texttt{tr}}|$ (\eg, 50+) and a large number of computation iterations~(\eg, 1000+) for a single batch~\cite{grosse2017statistical,rabanser2018failing,gao2020maximum,jia2021efficient}. 

\section{Our approach}

An overview of our approach is illustrated in \cref{fig:pipeline}. Our key idea is that, instead of constructing additional specialized witness function, one can instantiate the witness function using the off-the-shelf model pre-trained on the training data $\mathcal{D}_\texttt{tr}$. This witness function leads to the proposed metric, Neural Mean Discrepancy (\method), which evaluates the statistics of neural activations from the off-the-shelf model. 

\subsection{Neural Mean Discrepancy}
\label{sec:method_define}
The supremum in~\cref{equ:ipm_emp} is taken over the witness function $\varphi(\cdot)$, which implies that the neural network $\varphi_\omega(\cdot)$ is optimized to maximize the discrepancy in expectation over $\mathbb{Q}$ and $\mathbb{P}$~\cite{li2017mmd,binkowski2018demystifying,liu2020learning}. This optimization leads to high computational cost. Instead, we propose to relax the requirement of supremum in the context of OOD detection by making an intuitive assumption: as long as a function is capable of differentiating the statistics (\ie, mean) of examples from in- and out-of- distributions in the projected (\ie, feature) space, this function can be a qualified witness function. Interestingly, we find the off-the-shelf model $f(\cdot)$ pre-trained on the in-distribution training set fits this criteria. 

Taking a certain channel $c$ in the $l$-th layer of the off-the-shelf model as function $f_c^l: \mathbb{R}^{|\mathcal{I}|\times3\times d'\times d'}\rightarrow\mathbb{R}^{|\mathcal{I}|\times1\times d\times d}$, where $d'$ and $d$ are the spatial sizes of input images and activation maps, respectively. We define a model-agnostic metric named Neural Mean Discrepancy (\method) using $f_c^l$ as the witness function, 
\begin{align}
    \texttt{\method}_c^l(\mathcal{I}) =&\  \frac{1}{|\mathcal{I}|\cdot d^2}\sum_{i=1}^{|\mathcal{I}|}\sum_{m=1}^{d}\sum_{n=1}^{d} f^l_c(\rvx_i)_{m,n} \label{eq:input_mean}\\
    -&\  \frac{1}{|\mathcal{D}_\texttt{tr}|\cdot d^2}\sum_{j=1}^{|\mathcal{D}_\texttt{tr}|}\sum_{m=1}^{d}\sum_{n=1}^{d} f^l_c(\rvs_j)_{m,n}\label{eq:trainset_mean}\\
    =&\ \boldsymbol{\mu}[f^l_c(\mathcal{I})] - \boldsymbol{\mu}[f^l_c(\mathcal{D}_\texttt{tr})]~,
    \label{eq:short_fmd}
\end{align}
where the first sum ($ \sum_{i=1}^{|\mathcal{I}|}$ or $\sum_{i=1}^{|\mathcal{D}_\texttt{tr}|}$) is taken over examples and the last two sums ($\sum_{m=1}^{d}\sum_{n=1}^{d}$) are taken over all spatial positions $(m,n)$ in this channel. 

We sum over spatial positions of the activation map because each kernel of a neural net can be viewed as a realization of the witness function in the IPM theory.  Thus taking average over spatial positions within a channel (\textit{i.e.}, output of a kernel) is a faithful implementation of IPM with an NN. 

Each spatial position responds to a corresponding patch from the input image, known as the \textit{receptive field}~\cite{matsugu2003subject,luo2016understanding,araujo2019computing}. As a result, averaging across spatial positions can be thought of as averaging over image patches after projecting them with $f_c^l$. This implicitly augments the input batch and enables our method to survive from an extremely small batch size $|\mathcal{I}|$ (even for a single input image $|\mathcal{I}|=1$) when compared to previous IPMs-based methods.

\paragraph{Multi-layer \method for multi-scale OOD detection.}
To further improve the performance, we consider measuring and combining {\method}s from all channels across layers in the off-the-shelf model. By doing that, we can get an \method vector for a given input batch $\mathcal{I}$,
\begin{equation} 
    \resizebox{\linewidth}{!}{$
    \texttt{\method}(\mathcal{I}) = \{\texttt{\method}_1^1, \texttt{\method}_2^1, \dots, \texttt{\method}_1^l, \texttt{\method}_2^l, \dots, \texttt{\method}_1^L, \texttt{\method}_2^L, \dots\}$}
    \label{eq:concat_fmd}
\end{equation} which is a $C$-dimensional vector where $C$ is the total number of channels in the off-the-shelf model. 
The multi-layer {\method} has three major advantages:
\begin{enumerate}
\itemsep -2pt
    \item Each \methodlong $\texttt{\method}_c^l$ associates with a unique witness function $f_c^l$. Our method utilises the combination of several witness functions which deliver richer capacity than previous approaches based on a single IPM, as validated by our extensive experiments. 
    \item $\texttt{\method}_c^l$ for different layers may have different patch sizes because their \textit{receptive fields} increase linearly with their layer depths. Combining {\method}s from all layers enable multi-scale OOD detection which captures both low-level and high-level semantics (See \cref{sec:ablation}).
    \item By using multiple channels, \method does not introduce extra computation overhead since they can be obtained via a single forward pass of the model.
\end{enumerate}

\paragraph{``Free lunch'' from Batch Normalization.}

The way that \method computes the activation statistics coincides with what Batch Normalization (BN) does. Rather than computing $\boldsymbol{\mu}[f^l_c(\mathcal{D}_\texttt{tr})]$ by traversing the entire training data in~\cref{eq:trainset_mean}, one can directly use the running average from BN.

BN is an indispensable component in modern DNNs due to its ability of stabilizing training and improving model generalizability~\cite{ioffe2015batch}.
BN computes an output which normalizes input using per-channel statistics. 
Concretely, in a given channel, BN subtracts the activation mean $\mu$ from the inputs and then divides them by standard deviation $\sigma$. 
During training, $\mu$ and $\sigma^2$ are the empirical per-channel mean $\mu_\mathrm{batch}$ and variance $\sigma^2_\mathrm{batch}$ of the current mini-batch. During testing $\mu$ and $\sigma^2$ are not computed from mini-batches. Instead, the expected statistics \(\overline{\mu}\), $\overline{\sigma}^2$ are estimated from the training set and used for normalization. Ioffe \etal~\cite{ioffe2015batch} proposes that running average can be used to efficiently estimate expected statistics, 
\begin{equation}
    \resizebox{0.9\linewidth}{!}{$\overline{\mu} \leftarrow \lambda\overline{\mu} + (1-\lambda)\mu_\mathrm{batch},\quad \overline{\sigma}^2\leftarrow \lambda\overline{\sigma}^2 + (1-\lambda)\sigma^2_\mathrm{batch},$
    \label{equ:running_mean}}
\end{equation}
where a typical value of $\lambda$ is $0.99$ (which is a standard way of implementation in most deep learning libraries~\cite{paszke2019pytorch,abadi2016tensorflow}).

Back to our method, we use the running average mean $\overline{\mu}$ stored in BN directly to approximate $\boldsymbol{\mu}[f^l_c(\mathcal{D}_\texttt{tr})]$ instead of manually computing it with~\cref{eq:trainset_mean},
\begin{equation}
    \boldsymbol{\mu}[f^l_c(\mathcal{D}_\texttt{tr})] \approx \overline{\mu}^\ell_c. 
    \label{equ:approx_with_bn}
\end{equation}
We adapt this approximation in our experiments and validate that it works efficiently for OOD detection. Besides, we also validate the effectiveness of \cref{eq:trainset_mean} for models not containing BN (\eg, VGG~\cite{simonyan2014very} and Transformer~\cite{dosovitskiy2020vit}).

\subsection{A proof of concept}
\label{sec:our-avg}
To verify our intuition using an example, we instantiate the in-distribution data using CIFAR-10~\cite{krizhevsky2009learning} and the out-of-distribution data using SVHN~\cite{netzer2011reading}. A ResNet-34~\cite{he2016deep} is trained on CIFAR-10 with standard training receipt as the off-the-shelf model $f(\cdot)$. 
Given a mini-batch $\mathcal{I}$, 
its \method vector is computed via~\cref{eq:short_fmd,eq:concat_fmd,equ:approx_with_bn}.
We propose an intuitive baseline method called \textit{Ours-Avg}, which takes the average over elements' magnitude in the \method vector as confidence score for OOD detection.  
We randomly sample 100 mini-batches from CIFAR-10~(\textcolor{mygreen}{green dots}) and SVHN~(\textcolor{myred}{red dots}) testing sets and visualize each batch's score in~\cref{fig:scatter}. The observation in~\cref{fig:scatter} validates our expectation: OOD data has larger \method than in-distribution data on average.

Without any training, model fine-tuning, or hyper-parameter tuning, \textit{Ours-Avg} achieves an impressive performance, 99.9\% AUROC, with batch size $|\mathcal{I}|=4$. In contrast, other IPM-based methods typically require the batch size to be much larger~\cite{grosse2017statistical,rabanser2018failing,gao2020maximum}.

\subsection{A sensitivity-aware \method detector}
\label{sec:method_train_ood}

To further improve discriminatory power of the OOD detection, we propose to learn a parametric detector that takes \method vectors as input instead of simply averaging them. By doing this, the detection performance is boosted even the batch size $|\mathcal{I}|$ drops to 1 (\ie, single input example).

Previous literature~\cite{han2015learning,dong2017learning,bau2017network} observed that channels in deep neural networks are correlated and of different importance. To leverage this observation, we propose to train a detector $g(\cdot)$ that takes the \method vector $\texttt{\method}(\rvx)$ as input and predicts whether the current example is OOD or not. During training, these detectors are optimized on pairs of \method representations and distribution indicators, \eg, $(\texttt{\method}(\rvx_\mathrm{ID}), 0)$ for in-distribution examples and $(\texttt{\method}(\rvx_\mathrm{OOD}), 1)$ for out-of-distribution ones.

These OOD detectors are simple, lightweighted, and efficient at both training and inference. 
We will demonstrate, in the experimental section, that the detector can learn with few-shot examples and has high generalizability to unseen OOD types.
Even without access to OOD examples, the detector can still achieves superior performance by randomly permuting the pixels of in-distribution examples~\cite{ren2019likelihood}.

While the detector $g(\cdot)$ can be implemented using any classification method, we compare in our experiments two kinds of lightweight OOD detectors: a logistic regression $g_\texttt{LR}$~(LR) and a multilayer perceptron $g_\texttt{MLP}$~(MLP).
The whole pipeline of our method can be found in~\cref{alg:pipeline}.


\begin{algorithm}[!t]
\small
\caption{Pipeline of our NMD-based OOD detection}
\label{alg:pipeline}
\begin{algorithmic}
\Statex{{\bf \textit{Input}:} (1) an input example $\rvx$,

\quad(2) an off-the-shelf pre-trained classifier $f(\cdot)$, and

\quad(3) an OOD detector ($g_\texttt{LR}$ or $g_\texttt{MLP}$).
}
\vspace{0.1in}
\State \textit{{Stage 1: Generate feature mean discrepancy vector}} 
    \Indent
    \State Do a forward pass with the off-the-shelf model $f(\rvx)$ 
    \For{each channel in $f$}
    \State Compute $\texttt{\method}^l_c(\rvx)$ via~\cref{eq:short_fmd,eq:concat_fmd,equ:approx_with_bn}
    \EndFor
    \State $\texttt{\method}(\rvx)$ $\leftarrow$ Concatenate all channels' $\texttt{\method}^l_c(\rvx)$
    \EndIndent
\vspace{0.07in}
\Statex \textit{{Stage 2: Detect with the generated \method vector}} $\texttt{\method}(\rvx)$
    \Indent
    \If{Training}
    \State Train the OOD detector $g(\cdot)$ with pairs:
    
    \quad\ \ $\{\ (\texttt{\method}(\rvx_\mathrm{ID}), 0),\ \dots,\  (\texttt{\method}(\rvx_\mathrm{OOD}), 1)\ \}$
    \ElsIf{Testing}
    \State Use the OOD detector $g\left(\texttt{\method}\left(\rvx\right)\right)$ to get 
    
    \quad the detecting result
    \EndIf
    \EndIndent
\end{algorithmic}
\end{algorithm}


\section{Experimental setup}
\label{sec:exp_setup}

\fakeparagraph{Off-the-shelf models.} 
\method is model-agnostic and we evaluate it on multiple architectures, including 4-layer ConvNet~\cite{radfordUnsupervisedRepresentationLearning2016,jia2021efficient}, ResNet-34~\cite{he2016deep}, self-supervised ResNet-34~\cite{he2020momentum}, WideResNet~\cite{zagoruyko2016wide}, DenseNet-100~\cite{huang2017densely}, VGG~\cite{simonyan2014very}, and Vision Transformer~\cite{dosovitskiy2020vit}. 
All models are well-trained using their original training receipts and frozen (\ie, no fine-tuning) throughout the experiments. 

\fakeparagraph{Benchmark datasets.} 
    We perform comparative studies on various datasets: CIFAR-10, CIFAR-100, SVHN, cropped ImageNet, cropped LSUN, iSUN, and Texture, following OOD literature~\cite{liang2017enhancing,lee2018simple,ren2019likelihood,liu2020energy,sastry2020detecting}.
    Different combinations of in- and out-of- distribution datasets result in different levels of difficulty. An OOD detection problem is typically categorized into \textit{near}-OOD and \textit{far}-OOD \cite{winkens2020contrastive,fort2021exploring,ren2021simple}. \textit{Near}-OOD means that the two data distributions are close to each other. An example is using CIFAR-10 as in distribution and CIFAR-100 as OOD. This is because both datasets come from the same tinyimagenet dataset \cite{prabhu2020large} and their labels are all daily objects with similar semantics. In contrast, an example for \textit{far}-OOD could be CIFAR-10 as in distribution and SVHN as OOD because SVHN contains only house number images while CIFAR-10 contains natural images with rich information. \textit{Near}-OOD is generally a harder task than \textit{far}-OOD~\cite{sastry2020detecting,zhang2020hybrid,ren2019likelihood}. 
    In order to demonstrate the effectiveness of our approach, we evaluate the \method method in both \textit{near-}OOD and \textit{far-}OOD tasks.

\fakeparagraph{Protocols.}
We consider 4 kinds of data access circumstances to simulate real-world OOD detection scenarios. 
\begin{enumerate}
\itemsep -3pt
    \item \textit{Full access}: Conventional OOD detection approaches assume the access to both ID and OOD data for OOD detector training and hyper-parameter tuning. 
    \item \textit{Few-shot}: Due to privacy concerns, the data owner may only release a few ID and OOD showcase examples for OOD detector training. In our experiments, we propose an extreme scenario where one only has access to 25 ID and 25 OOD examples for training.  
    \item \textit{Zero-shot}: Recent studies~\cite{hsu2020generalized,liu2020energy,sastry2020detecting,zaeemzadeh2021ood} also learn OOD detectors with only ID examples and without any dependence on OOD examples.
    \item \textit{Transfer}: To evaluate the transferability of different methods, we additionally propose to train the detectors on one kind of OOD dataset and evaluate their performance on separate unseen OOD datasets. 
\end{enumerate}

\fakeparagraph{Evaluation metrics.} Consistent with the literature~\cite{liang2017enhancing,lee2018simple,ren2019likelihood,liu2020energy,sastry2020detecting}, we use three evaluation metrics: (1) true
negative rate at 95\% true positive rate (TNR95), (2) area
under the receiver operating characteristic curve (AUROC),
and (3) detection accuracy (ACC) which measures the maximum detection accuracy over all possible thresholds.

\fakeparagraph{Baseline methods.} 
We compare our approach with several existing methods lying in different categories.
\begin{enumerate}
\itemsep -3pt
    \item \textit{Statistical methods}: These are most related to our work. As summarized in Sheng~\etal~\cite{jia2021efficient}, for a test example $\rvx$, such approaches compute the OOD score using the negative of the sum of kernel evaluation at each of the inlier example $\mathcal{S}_{\rvx'}$ such that $\textsc{score}(\rvx)=-\sum^{|\mathcal{S}|}_{i=1}\kappa(\rvx, \rvx'_i)$. Different choices of the kernel $\kappa$ result in different methods including Deep kernel~(DK~\cite{liu2020learning,gao2020maximum}), convolution neural tangent kernel~(CNTK~\cite{arora2019exact}), and shift-invariant convolutional
neural tangent kernel~(SCNTK~\cite{jia2021efficient}).

    \item \emph{Other baselines:} We also compare our method with other state-of-the-art approaches such as ODIN~\cite{liang2017enhancing}, Mahalanobis distance~\cite{lee2018simple}, OE with classifier fine-tuning~\cite{hendrycks2018deep}, and Energy with classifier fine-tuning~\cite{liu2020energy}. They require model fine-tuning, hyper-parameter tuning, multi-round forward inference, while \method does not depend on any of the above.
    
\end{enumerate}

\section{Results}
We show our results in this section, which empirically demonstrate the simplicity, efficacy, efficiency, and generalizability of \method-based OOD detection. All results are obtained for single example detection, \ie, batch size $|\mathcal{I}=1|$.

\subsection{Comparison with statistical baselines}
\label{sec:comp_stat_method}

We first compare our method with the most related line of approaches based on statistical tests, \ie, DK~\cite{liu2020learning,gao2020maximum}, CNTK~\cite{arora2019exact}, and SCNTK~\cite{jia2021efficient}. These methods require a traversing in a subset of the in-distribution data $\mathcal{S}_{\rvx'}$ for every test example, which could be expensive. \method does not depend on $\mathcal{S}_{\rvx'}$ which leads to higher efficiency. Following settings of Sheng~\etal~\cite{jia2021efficient}, all compared methods adapt a four-layer convolutional neural network as the feature extractor. \cref{tab:stat_method} shows that our method~(using logistic regression detection, denoted as `\textit{Ours-LR}') achieves significantly better OOD detection performance (99.8+\% AUROC). The result empirically  justifies the value of using multiple witness functions at different scales and semantic levels from the same pre-trained model.

\begin{table}[t]
\small
\centering
\resizebox{\linewidth}{!}{
\begin{tabular}{cccc|c}
\toprule
    \begin{tabular}[c]{@{}c@{}} Model \end{tabular} &
    \begin{tabular}[c]{@{}c@{}} ID \end{tabular} &
    \begin{tabular}[c]{@{}c@{}} OOD \end{tabular} &
    \begin{tabular}[c]{@{}c@{}} Method \end{tabular} &
    AUROC\\                
\midrule
\multirow{4}{*}{ \begin{tabular}[c]{@{}c@{}} ConvNet \\ (4 layers)\end{tabular} } & 
\multirow{4}{*}{ \begin{tabular}[c]{@{}c@{}} CIFAR-10 \end{tabular} } & 
\multirow{4}{*}{SVHN} & 
{DK} & 82.4\\
&&&{CNTK}& 71.3\\
&&&{SCNTK}& 84.9\\
&&&{Ours-LR}& \textbf{99.9}\\
\midrule
\multirow{4}{*}{ \begin{tabular}[c]{@{}c@{}} ConvNet \\ (4 layers)\end{tabular} } & 
\multirow{4}{*}{ \begin{tabular}[c]{@{}c@{}} SVHN \end{tabular} } & 
\multirow{4}{*}{CIFAR-10} & 
{DK} & 21.4 \\
&&&{CNTK}& 51.9\\
&&&{SCNTK}& 80.3\\
&&&{Ours-LR}& \textbf{99.8}\\
\bottomrule
\end{tabular}}
\caption{AUROC comparison of \textbf{statistical OOD detection} methods. we compare our method with deep kernel on extracted feature maps (DK)~\cite{liu2020learning,gao2020maximum}, convolutional neural tangent kernel (CNTK)~\cite{arora2019exact}, shift-invariant convolutional neural
tangent kernel~(SCNTK)~\cite{jia2021efficient}. Consistent to the setting of \cite{jia2021efficient}, we use a four-layer convolution neural network as the classifier for feature extraction. More details can be found in the appendix.}
\label{tab:stat_method}
\vspace{-3mm}
\end{table}

\subsection{Comparison with other baselines}

\begin{figure*}[t]
    \centering
    \includegraphics[width=\textwidth]{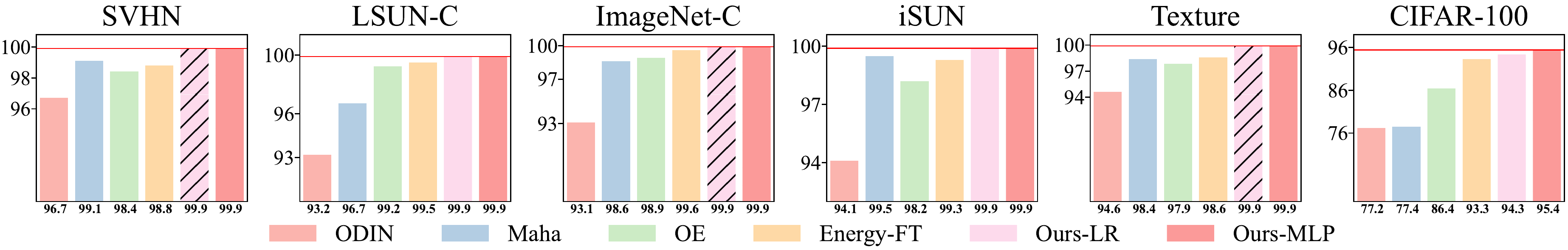}
    \caption{AUROC comparison with OOD methods requiring \textbf{both in- and out-of- distribution data} for detector training, classifier fine-tuning or hyper-parameter search. We compare our method with ODIN~\cite{liang2017enhancing}, Maha~\cite{lee2018simple}, OE with  fine-tuning~\cite{hendrycks2018deep}, and Energy with fine-tuning~\cite{liu2020energy} on ResNet-34, using CIFAR-10 as in-distribution. 
    The Energy method is finetuned using each of the OOD training sets.}
    \label{fig:compare_in_out_data}
\end{figure*}

We evaluate our method on a set of out-of-distribution datasets with ResNet trained on the in-distribution dataset CIFAR-10. In this experiment, we assume both the ID and OOD datasets are available for training. The pre-trained ResNet-34 is frozen in our \method method, while other methods may further fine-tune it to maximize the test power. In addition, our \method is hyper-parameter free while other approaches may have sensitive hyper-parameters to tune~(e.g., temperatures in \cite{liang2017enhancing}, perturbation in \cite{lee2018simple}, and margin in \cite{liu2020energy}). As shown in~\cref{fig:compare_in_out_data}, despite its simplicity, our method consistently outperforms other methods across datasets, especially on \textit{near-}OOD dataset, CIFAR-100. More experimental results can be found in the appendix.

\begin{figure*}[t]
    \centering
    \includegraphics[width=\textwidth]{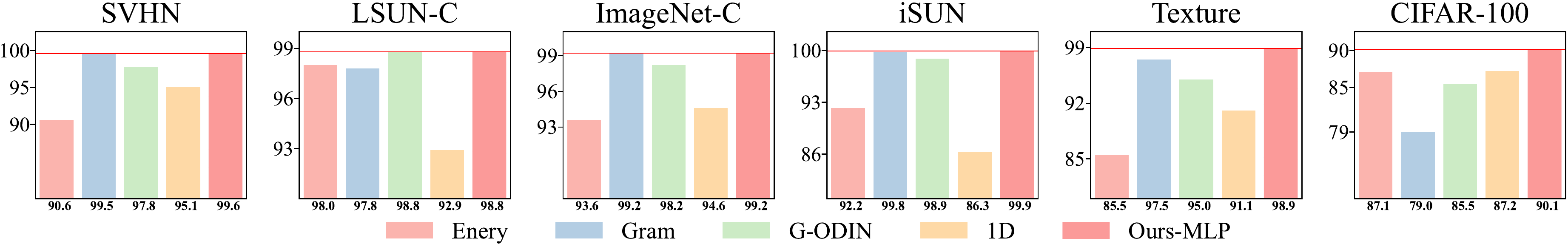}
    \caption{AUROC comparison of detection methods when \textbf{only in-distribution dataset is accessible}. We compared our method with Energy without classifier fine-tuning~\cite{liu2020energy}, Gram~\cite{sastry2020detecting}, G-ODIN~\cite{hsu2020generalized}, and 1D~\cite{zaeemzadeh2021ood} on ResNet-34, using CIFAR-10 as in-distribution dataset.}
    \label{fig:compare_only_in}
\end{figure*}

\begin{figure*}[t]
    \centering
    \includegraphics[width=\textwidth]{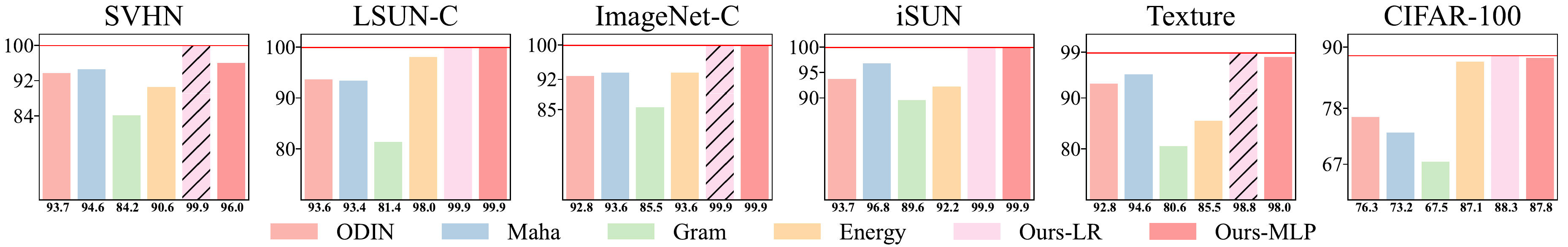}
    \caption{AUROC comparison of detection methods when \textbf{only 25 in- and 25 out-of- distribution examples are accessible}. We compared our method with ODIN, Maha, Gram, and Energy with classifier fine-tuning~\cite{liu2020energy} on ResNet-34, using CIFAR-10 as in-distribution dataset.}
    \label{fig:compare_25in_25out}
\end{figure*}

\subsection{Learning with only in-distribution examples}
\label{sec:no_ood}
We further compare our method with approaches that do not depend on any given OOD dataset for training. Among them, G-ODIN~\cite{hsu2020generalized} and 1D~\cite{zaeemzadeh2021ood} need to fine-tune the model on in-distribution dataset. Since no OOD example is accessible, we craft artificial OOD examples by randomly permuting pixels of in-distribution examples and use the crafted OOD examples to train our detector. The detector trained on artificial OOD examples is evaluated on realistic OOD datasets. 
\cref{fig:compare_only_in} shows that our method performs better than state-of-the-art without access to realistic OOD data. The result also suggests that, even though the artificial OOD examples are unrealistic, they are helpful in guiding the decision boundary of an OOD detector.

\subsection{Few-shot OOD training}
We evaluate our method under the scenario that a very limited number of in-distribution and out-of-distribution examples are available for training. 

\cref{fig:compare_25in_25out} compares different methods when only 25 ID examples and 25 OOD examples are present during training. The baseline `Gram' uses 50 ID examples as an exception because it does not depend on OOD examples. Since 50 examples are too few to conduct fine-tuning for `Energy', we report its performance without fine-tuning as a reference. 

Our method outperforms all other methods under this few-shot setting. Previous works often require sufficient data to tune hyper-parameters or models. In contrast, \method is hyper-parameter-free and thus can learn well with few examples. However, we observe a slight over-fitting of the MLP detector which suggests one should consider low-capacity models such as LR in the few-shot cases.

\begin{table*}[!h]
\centering
\resizebox{\linewidth}{!}{
\begin{tabular}{ccc|ccccc}
\toprule
    \multirow{2}{*}{\begin{tabular}[c]{@{}c@{}} In-dist. \end{tabular}} & \multirow{2}{*}{\begin{tabular}[c]{@{}c@{}} Train \\ OOD \end{tabular}} &
    \multirow{2}{*}{\begin{tabular}[c]{@{}c@{}} Test \\ OOD \end{tabular}} & \textbf{ResNet-34}  & \textbf{ResNet-34 (self)} & \textbf{VGG-19} & \textbf{ViT} & \textbf{DenseNet}\\ 
     \cline{4-8} 
    &   &  & \multicolumn{5}{c}{ {\scriptsize TNR at TPR 95\% $\uparrow$ / AUROC $\uparrow$ / ACC $\uparrow$ \par} }                \\ 
\midrule
\multirow{4}{*}{\begin{tabular}[c]{@{}c@{}} {\footnotesize CIFAR-10} \end{tabular}} & \multirow{4}{*}{\begin{tabular}[c]{@{}c@{}} {\footnotesize CIFAR-100} \end{tabular}}
& LSUN-C
& \multicolumn{1}{c}{95.8 / 99.2 / 95.6}
& \multicolumn{1}{c}{99.1 / 99.8 / 98.1}
& \multicolumn{1}{c}{96.4 / 99.3 / 95.7}
& \multicolumn{1}{c}{94.0 / 98.7 / 94.6}
& \multicolumn{1}{c}{90.6 / 98.3 / 93.6}
\\
& & SVHN
& \multicolumn{1}{c}{96.4 / 99.2 / 95.9}
& \multicolumn{1}{c}{99.9 / 99.9 / 99.9} 
& \multicolumn{1}{c}{99.9 / 99.9 / 99.1} 
& \multicolumn{1}{c}{99.8 / 99.9 / 99.2}
& \multicolumn{1}{c}{95.8 / 99.2 / 95.4}
\\
& & Texture
& \multicolumn{1}{c}{91.7 / 98.5 / 93.4}
& \multicolumn{1}{c}{97.8 / 99.5 / 96.7} 
& \multicolumn{1}{c}{96.1 / 99.1 / 95.6} 
& \multicolumn{1}{c}{91.4 / 98.3 / 93.5}
& \multicolumn{1}{c}{93.0 / 98.6 / 94.0}
\\
& & ImageNet-C
& \multicolumn{1}{c}{93.7 / 98.7 / 94.4}
& \multicolumn{1}{c}{99.9 / 99.9 / 99.1} 
& \multicolumn{1}{c}{94.0 / 98.9 / 94.5} 
& \multicolumn{1}{c}{89.0 / 98.1 / 93.0}
& \multicolumn{1}{c}{94.3 / 98.8 / 94.7}
\\
\bottomrule
\end{tabular}}
\caption{We evaluate generalizability of our method across models, including ResNet-34 trained with standard softmax cross-entropy loss~\cite{he2016deep}, ResNet-34 trained with self-supervised loss from MoCo~\cite{he2020momentum}, VGG-19~(without BN)~\cite{simonyan2014very}, and Visual Transformer~\cite{dosovitskiy2020vit}. To further validate the generalizability across datasets, we use CIFAR-100 as OOD dataset to train our detector and test the trained detector on \textbf{unseen OOD datasets} including LSUN-C, SVHN, Texture, and ImageNet-C.}
\label{tab:more_models_transfer}
\vspace{-3mm}
\end{table*}



\subsection{Generalizability across models and datasets}
\label{sec:exp_model_nobn}

We are interested in the transferability of the detector across datasets. For each model, we use CIFAR-100 as OOD dataset for training the detector and evaluate the trained detector on unseen OOD datasets such as LSUN-C, SVHN, Texture, and ImageNet-C. 

As we elaborated in~\cref{sec:method_define}, one can either use running average in BN to approximate $\boldsymbol{\mu}[f^l_c(\mathcal{D}_\texttt{tr})]$~(\cref{equ:approx_with_bn}) or manually compute it via \cref{eq:trainset_mean} if the model has no BN layers. So we also evaluates the generalizability of our method across different models.

\begin{enumerate}
\itemsep -3pt

\item \textit{VGG models.} VGG-19 consists of 16 convolution and ReLU layers, followed by
three fully-connected (FC) layers. It has no BN layers. We only use channels from convolutional layers to compute \method. Since no BN layer is present in this model, we traverse the in-distribution training set (\ie, CIFAR-10) for one epoch.

\item \textit{Self-supervised models.} We use MoCo~\cite{he2020momentum} as the self-supervised learning method. After pre-training a ResNet-34 model with MoCo on CIFAR-10, we freeze it and use it to compute \method. 

\item \textit{Vision Transformers.} Different from CNNs, a Vision Transformer (ViT)~\cite{dosovitskiy2020vit} is composed of a stack of standard multi-head self-attention and position-wise fully-connected layers.
ViT splits an image
into $p$ non-overlapped patches and provides the sequence of embeddings of these patches as an input to a Transformer. ViT adopts layer normalization (LN)~\cite{ba2016layer} to normalize each input example's activation $\rmZ^l\in\mathbb{R}^{p\times d}$. 
Imitating convolution neural networks, we compute ViT's feature mean for an input example $\rvx$ with $\boldsymbol{\mu}^l\left(\rvx\right)=\frac{1}{p}\sum_{i=1}^p \ermZ^l_{i}\in\Re^{d}$, and use it to compute \method metric.
\end{enumerate}

\cref{tab:more_models_transfer} indicates that \method generalizes well for various models and datasets. 
Interestingly, we find that self-supervised ResNet-34 has the best averaged detection performance across 4 unseen OOD datasets, suggesting the high transferability of its learnt representations~\cite{he2020momentum,chenBigSelfSupervisedModels2020,ericssonHowWellSelfSupervised2021}.

\subsection{Training and inference efficiency}
\label{sec:training_inference_exp}

In this section, we compare the training and inference costs of the proposed Ours-MLP with baselines in~\cref{fig:latency}. We measure the training and inference time on a machine with one NVIDIA GPU 1080 Ti and a Intel(R) Xeon(R) CPU E5-2650 v4 @ 2.20GHz.

\fakeparagraph{Training cost.} 
Since the detectors (\ie, LR and MLP) we used are lightweight, the training process can be done quickly (within 60 epochs with CIFAR-10~(ID) and CIFAR-100~(OOD) training datasets in ~\cref{fig:latency}). In addition, different from existing methods, \method does not have sensitive hyperparameters and thus does not have to repeat training process for multiple times to search the hyperparameters.

\fakeparagraph{Inference cost.} As illustrated in~\cref{alg:pipeline}, we only have to run a single forward pass with the pre-trained model to generate the \method vector. The generated \method vector will be then processed by a lightweight detector (\eg, Logistic regress or three-layer MLP as detailed in~\cref{sec:method_train_ood}). 
In contrast, other approaches, in addition to a standard forward pass (Baseline~\cite{hendrycks2016baseline} and ACET ~\cite{hein2019relu}), also require either: (1) extra forward and backward passes~\cite{hsu2020generalized,lee2018simple,liang2017enhancing}; (2) computing complicated properties (e.g., co-occurrences) \cite{sastry2020detecting,zaeemzadeh2021ood}.

\subsection{Ablation study}
\label{sec:ablation}

\fakeparagraph{Layer importance for OOD detection.}
\begin{figure}
    \centering
    \includegraphics[width=0.85\linewidth]{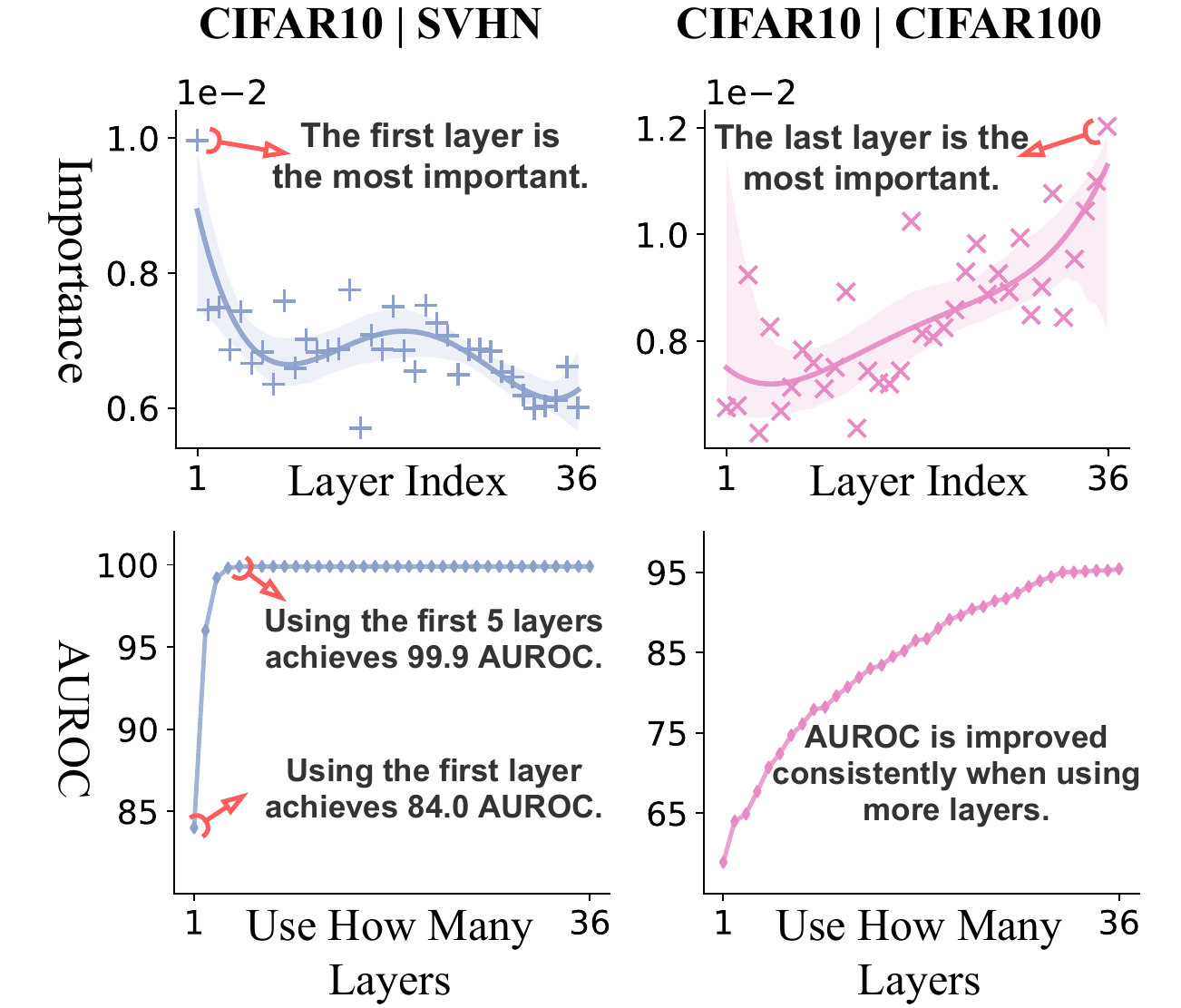}
    \caption{We study the layer-granularity importance of ResNet-34 for \textit{far-}OOD (\eg, SVHN, \textcolor{myblue}{\textbf{left}}) and \textit{near-}OOD~(\eg, CIFAR-100, \textcolor{mypink}{\textbf{right}}) using CIFAR-10 as in-distribution dataset. \textbf{Top figures}: Importance of each layer in ResNet-34, where a layer's importance is measured by the averaged magnitude of channels' coefficients in the LR detector. \textbf{Bottom figures}: We use the first $k$ layers~(x-axis) to do detection and evaluate the AUROC for different $k$.}
    \label{fig:layer_importance}
    \vspace{-4mm}
\end{figure}
We visualize the importance of \method from different layers of ResNet-34 in~\cref{fig:layer_importance}. We find that our method utilises both low-level visual attributes (from shallow layers) as well as high-level semantic information (from deep layers) and dynamically adjusts the importance depending on tasks. In the top two plots of~\cref{fig:layer_importance}, we specify the OOD detector as a logistic regression (LR). 
We standardize \method vector to ensure that each of its dimensions has a similar magnitude before sending it to the LR model.

The absolute values of each learned coefficient in LR can be treated as the importance of its corresponding channel.
We compute a layer's importance by averaging all its channels' importance values. 
We test two OOD detection tasks: (1) In a \emph{far}-OOD task CIFAR-10 against SVHN, {\method} values from shallow layers, which extract low-level features~\cite{10.5555/2969033.2969197,asano2019critical}, are already able to differentiate CIFAR-10 against SVHN. (2) In a \textit{near}-OOD task, CIFAR-10 against CIFAR-100, we have to rely more on \method values from deeper layers to capture the semantic differences. 

In the bottom two plots of~\cref{fig:layer_importance}, we visualize the detection performance when the first $k$ layers' {\method}s are used. For SVHN, it is sufficient to achieve the best AUROC with only the first 5 layers. While, for CIFAR-100, using more layers consistently leads to better performance. This observation aligns well with the previous paragraph, and motivates a potential future work on dynamically selecting layers given an input example (\ie, `early exits') for better performance and efficiency like~\cite{linMOODMultiLevelOutofDistribution2021}.      

\fakeparagraph{Does Neural Variance Discrepancy help?} Besides the first order statistics (\ie, mean), one can define Neural Variance Discrepancy (NVD) by computing the activation's second-order statistics in a similar manner. In practise, we find that using NVD (AUROC=95.3, CIFAR-10 against CIFAR-100 on ResNet-34) is able to achieve a similar OOD detection performance as NMD (AUROC=95.4). Combining both NMD and NVD obtains a slightly better result (AUROC=95.6). Please refer to \Cref{appendix:nvd} for more details.

\section{Related work}
\textit{OOD detection with model modification and fine-tuning.}
Multiple OOD approaches augment standard DNN architectures by 
model ensemble~\cite{gal2016dropout,lakshminarayanan2017simple,vyas2018out,chen2020oodanalyzer}, training an additional branch~\cite{devries2018learning,havasi2020training}, and learning a background model~\cite{ren2019likelihood}. In addition, various novel training objectives have been proposed including
training with OOD uniform label~\cite{lee2017training}, an additional OOD class~\cite{chen2020informative,mohseni2020self}, OOD examples crafted by a generative model~\cite{moller2021out}, energy score regualarization~\cite{liu2020energy}, and soft-binning error~\cite{karandikar2021soft}.

\emph{OOD detection without fine-tuning.}
Maximum softmax probability~\cite{hendrycks2016baseline} and it variants like ODIN~\cite{liang2017enhancing}, GODIN~\cite{hsu2020generalized}, POOD~\cite{tajwar2021no} and Energy~\cite{liu2020energy} have been used for OOD detction. Beside the final outputs, intermediate activation is also used like Gram~\cite{sastry2020detecting}, Mahalanobis~\cite{lee2018simple,ren2021simple}.

\emph{Statistical OOD detection.} Previous studies have shown some preliminary results of using Frechet distance~\cite{dowson1982frechet} and Maximum Mean Discrepancy (MMD)~\cite{gretton2012kernel} for adversarial~\cite{grosse2017statistical,carlini2017adversarial,gao2020maximum,roth2019odds} and distribution shift detection~\cite{rabanser2018failing,guillory2021predicting}.  
Erdil~\etal~\cite{erdil2021taskagnostic} applies adversarial perturbation and kernel density estimation~(KDE) for a subset of in-distribution examples and each input to do OOD detection. Density of states estimator is also used for OOD detection with generative models~\cite{morningstarDensityStatesEstimation2021}. Jia~\etal~\cite{jia2021efficient} proposes a compositional kernel, as a variates of adaptive deep neural kernel~\cite{liu2020learning,gao2020maximum,arora2019exact}, for efficient OOD detection. However, mere shallow models are considered. In this work, by creatively leveraging neural means from a pre-trained model, we significantly reduce the algorithmic complexity and computational cost of statistical OOD detection.

\emph{Updating Batch Normalization statistics for improved accuracy.}
Previous studies find that one can improve model performance by updating the statistics and affine parameters of the Batch Norm layers when either data or model change during training~\cite{tang2021crossnorm,rusak2021adapting} or testing~\cite{nado2020evaluating,you2021test}. 
Such BN recalibration techniques show promising effectiveness of improving the model performance~\cite{wu2021rethinking,hubara2021accurate}, domain adaptation and generalization ability \cite{li2018adaptive,salvador2021improved}, few-shot learning~\cite{du2020metanorm}, and the model's robustness against input noise~\cite{schneider2020improving,nandy2021adversarially,benz2021batch}.

\section{Conclusion and discussion}
\label{sec:conclusion}
We proposed Neural Mean Discrepancy (\method) which compares the neural means between test examples and training data for OOD detection. Both the IPMs-based theoretical analysis and empirical results validate the efficacy of \method. With the extreme algorithmic simplicity, \method is evaluated across datasets, models, and data access circumstances, achieving state-of-the-art accuracy and efficiency. 

\paragraph{Limitations.} Although \method can achieve competitive results without access to real OOD data~(see \cref{sec:no_ood}), artificially crafted OOD data via pixel shuffling is still required to learn channel sensitivities. It is likely to estimate sensitivities directly via weight distributions of the off-the-shelf model~\cite{yin2020dreaming,huang2020convolution} or gradient information~\cite{sundararajan2017axiomatic,adadi2018peeking}. In addition, early existing works~\cite{abdelzad2019detecting,linMOODMultiLevelOutofDistribution2021} could be applied to further improve the performance of an NMD-based OOD detector.

\clearpage
{\small
\bibliographystyle{ieee_fullname}
\bibliography{egbib}
}
\clearpage

\appendix

\section{Extra Results Using Different In-Distribution Datasets}
We further evaluate our method on various in-distribution datasets including SVHN and CIFAR-100 in~\cref{tab:more_dataset} using the same setting as~\cref{fig:compare_in_out_data}. According to \cref{fig:compare_in_out_data}, our method achieves state-of-the-art results with various in-distribution datasets.    

\begin{table}[!h]
\centering
\resizebox{0.48\textwidth}{!}{
\begin{tabular}{cccccc}
\toprule
     \multirow{1}{*}{\begin{tabular}[c]{@{}c@{}} In-dist. (model) \end{tabular}}  &   \multirow{1}{*}{OOD}      &    Baseline~\cite{hendrycks2016baseline} & ODIN~\cite{liang2017enhancing}  &  Maha.~\cite{lee2018simple} & Ours-LR \\ 
\midrule
\multirow{3}{*}{\begin{tabular}[c]{@{}c@{}} CIFAR-100 \\(ResNet-34) \end{tabular}}
&  SVHN &  79.5 & 70.7 & 92.4 & \textbf{94.2} \\
&  LSUN-C &  75.8 & 85.6 & 98.2 & \textbf{99.9} \\
&  ImageNet-C & 77.2 & 87.8 & 98.0 & \textbf{99.9} \\
\midrule
\multirow{3}{*}{\begin{tabular}[c]{@{}c@{}} SVHN \\(ResNet-34) \end{tabular}}
&  CIFAR-10 & 92.9 & 92.1 & 99.3 & \textbf{99.8} \\
&  LSUN-R & 91.6 & 89.4 & \textbf{99.9} & \textbf{99.9}  \\
&  ImageNet-R & 93.5 & 92.0 & \textbf{99.9} & \textbf{99.9} \\
\midrule
\multirow{3}{*}{\begin{tabular}[c]{@{}c@{}} CIFAR-100 \\(DenseNet-100) \end{tabular}}
&  SVHN &  82.7 & 85.2 & 90.3 & \textbf{93.3} \\
&  LSUN-C & 70.8 & 85.5 & 98.0 & \textbf{99.9} \\
&  ImageNet-C & 71.6 & 84.8 & 94.1 & \textbf{99.6} \\
\bottomrule
\end{tabular}
}
\caption{AUROC comparison of detection methods on various in-distribution datasets.}
\label{tab:more_dataset}
\end{table}

\begin{figure*}[h!]
    \centering
    \includegraphics[width=0.7\linewidth]{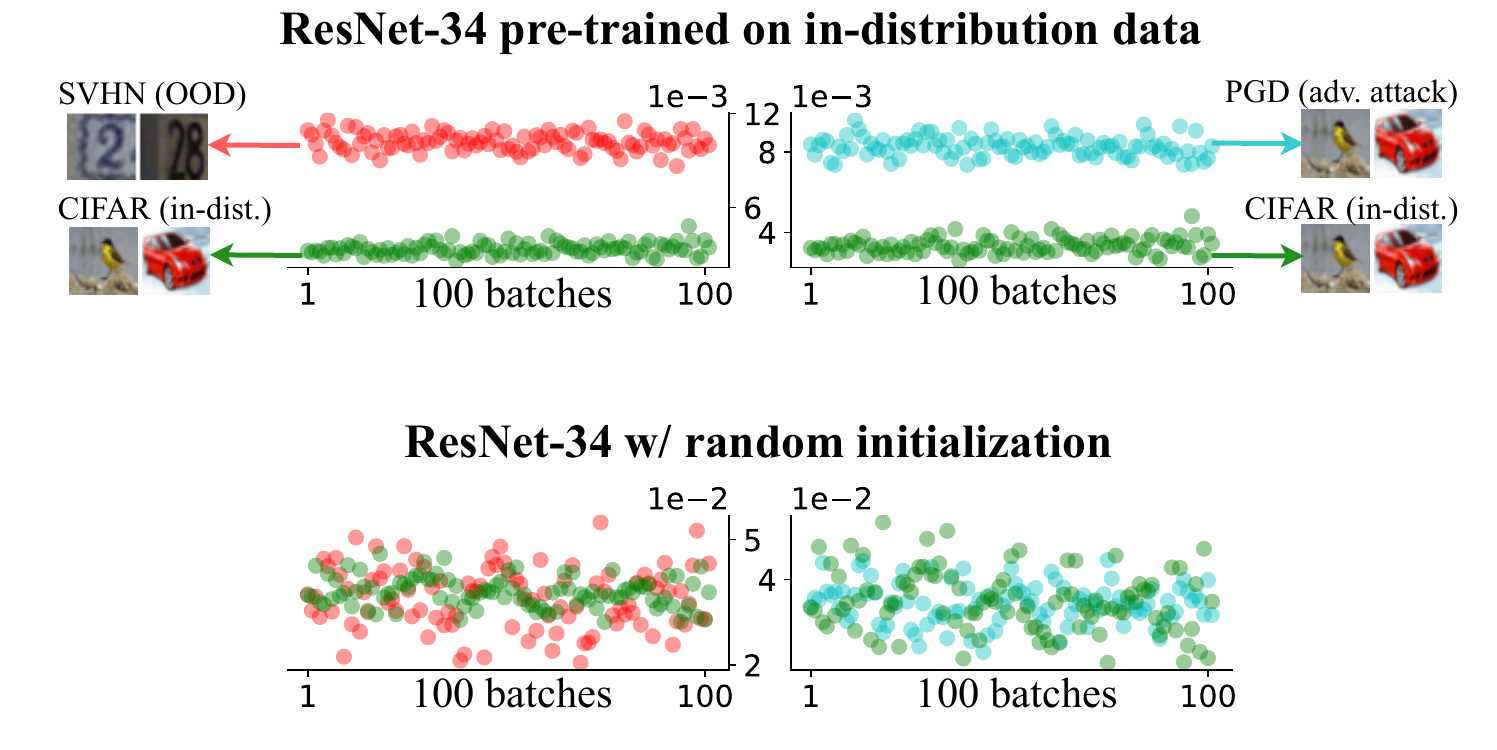}
    \caption{We redo the proof-of-concept experiment in~\cref{fig:scatter} with an un-trained ResNet-34. The batch size is 8. }
    \label{fig:random_scatter}
\end{figure*}

\section{Lightweight OOD Detector}
\begin{itemize}
\item \textbf{Logistic regression (LR)} is a kind of classic machine learning model for binary classification. Given an input vector, The LR model performs a dot product on the input with the learned coefficient vector and outputs the prediction score after applying the sigmoid function. A LR model can be trained efficiently by several solvers like LBFGS.~\cite{lee2006efficient} We use the default hyper-parameters in sklearn~\cite{sklearn.linear_model.LogisticRegressionCV} for training of LR detector.
\item The \textbf{multilayer perceptron (MLP)} we use consists of three full-connected layers following by non-linear activation function ReLU. We adapt dropout after the second full-connected layer and train the MLP with SGD optimizer. As a non-linear model, MLP is able to learn more complex correlation among elements in the input than LR. In practise, we find that MLP has slightly better detection performance than LR, while, with higher over-fitting possibility when the number of training examples is limited. We training the MLP using SGD with 0.001 learning rate and 0.9 momentum.   
\end{itemize}

The adapted OOD detector is lightweight. For instance, Our LR model has 8k parameters and 16k FLOPs, significantly smaller than the pre-trained model (\textit{e.g.}, ResNet-34 with $\sim$2${\times10^4}$k parameters and $\sim$2${\times10^6}$k FLOPs).

\section{Architecture of ConvNet}
Consistent to the setting of~\cite{jia2021efficient}, we use a simple ConvNet in~\cref{sec:comp_stat_method,tab:stat_method}. ConvNet's architecture is summarized in~\cref{tab:conv_arch}.

\begin{table}[h]
    \centering
    \begin{tabular}{c|c}
    \toprule
        Layer & Configuration \\
    \midrule
        Conv1 & (3, 300, kernel size=4, stride=1)\\
        Conv2 & (300, 300, kernel size=4, stride=2)\\
        Conv3 & (300, 300, kernel size=4, stride=2)\\
        Conv4 & (300, 300, kernel size=3, stride=2)\\
        AvgPool & (kernel size=2)\\
        FC & (300, 10)\\
    \bottomrule
    \end{tabular}
    \caption{Architecture of ConvNet following~\cite{jia2021efficient}. After each convolutional layer, batch normalization and ReLU layers are applied.}
    \label{tab:conv_arch}
\end{table}

\section{Pre-trained or Un-trained Models?}
\label{append:random_network}
In~\cref{fig:scatter}, we show that the average of elements' magnitude in \method vector from a pre-trained ResNet-34 can be used as OOD score to reliably distinguish OOD batches. Such a proof-of-concept example validates that the off-shelf-shelf pre-trained model can be used as a qualified witness function. Based on this interesting and supervising finding, we believe the off-the-shelf model itself should contain sufficient information about the training data distribution because it was trained to capture training data's features. 

To further validate our hypothesis, we replace the pre-trained ResNet-34 with an un-trained ResNet-34 and re-run the experiment. As shown in~\cref{fig:random_scatter}, an un-trained ResNet-34 cannot act as a qualified witness function to detect OOD batches even the batch size is 8.

\section{Neural Variance Discrepancy}
\label{appendix:nvd}
As mentioned in~\cref{sec:ablation}, one can define Neural Variance Discrepancy (NVD) by computing the activation’s second-order statistics in a similar manner as \method, 

\begin{align}
    \texttt{NVD}_c^l(\mathcal{I}) =\sqrt{\boldsymbol{\sigma^2}[f^l_c(\mathcal{I})]} - \sqrt{\boldsymbol{\sigma^2}[f^l_c(\mathcal{D}_\texttt{tr})]}~,
    \label{eq:short_nvd}
\end{align}
where the second term can be approximated by BN's running average variance. Interestingly, NVD-based detection~(\ie, NVD-MLP) achieves a comparable detection performance as NMD. 

We further combine NVD and NMD via concatenating them together. Since elements in NVD and NMD may have different magnitude, we adopt the \texttt{standardizer} from \texttt{sklearn} to remove the mean and scale to unit variance for each dimension of NVM and NMD vectors before concatenating. Combining NMD and NVD obtains a slightly better detection result although extra computation overhead is introduced.

\section{Crafting OOD Data by Pixel Permuting}

As discussed in~\cref{sec:no_ood}, if no OOD example is accessible, we craft artificial OOD examples by  randomly permuting pixels of in-distribution examples and use  the crafted OOD examples to guide our detector for finding the decision boundary. The premise of using crafted OOD example is that the method has high generalizability across datasets (\ie, for unseen OOD data) as validated in~\cref{sec:exp_model_nobn}. Specifically, we do pixel permuting in the block granularity instead of in the pixel granularity~\cite{ren2019likelihood} to avoid tuning the hyperparameter ``mutation rate''. Taking CIFAR-10 example as an example, we split an image into 16 non-overlapping ($8\times8$) blocks and randomly permute their positions. Results of detection performance without OOD examples are shown in~\cref{fig:compare_only_in,tab:no_ood_data}.

\section{Training and Inference Efficiency}
In~\cref{sec:training_inference_exp}, we compare the training and inference costs of the proposed Ours-MLP with baselines as shown in~\cref{fig:latency}. Training and inference time are measured on a machine with one NVIDIA GPU 1080 Ti and a Intel(R) Xeon(R) CPU E5-2650 v4 @ 2.20GHz. Some approaches conduct model fine-tuning using MIT 80 Million Tiny Images Dataset which is not available any more. 
For those methods, we use the target OOD dataset (\ie, CIFAR-100 training set) to do fine-tuning but with the same number of iterations as using MIT 80 Million Tiny Images Dataset.
For methods which require repeating experiments for several times to search hyper-parameters, we count all such time into training time. 
To measure the inference latency, we repeat single example detection for 10,000 times and compute the average inference time for a single example. 

A recent study, MOOD~\cite{linMOODMultiLevelOutofDistribution2021}, achieves state-of-the-art inference efficiency leveraging early exiting~\cite{abdelzad2019detecting}.  We do not include MOOD in~\cref{tab:latency} because it depends a special architecture  with dynamic exits. In addition, our method is orthogonal to MOOD and could be combined for a future work as discussed in~\cref{sec:ablation,sec:conclusion}.

\begin{table}[]
    \centering
    \begin{tabular}{c|cll}
    \toprule
    Method & Fine-tuning & Training & Inference \\
    \midrule
    Gram & True & 330s & 0.37s \\
    Maha & False & 1397s & 25.7ms\\
    ODIN & False & 1270s & 16.0ms \\
    G-ODIN & True & 1830s & 22.1ms \\
    OE & True & 560s & 6.72ms \\
    GOOD & True & 756m & 47.4ms \\
    ACET & True & 201m & 6.89ms \\
    Energy-FT & True & 620s & 7.24ms\\ 
    \midrule
    \midrule
      Plain ResNet-34 & - &  - & 6.72ms \\
    \midrule
      Ours-MLP & False & 94s & 7.54ms \\
    \bottomrule
    \end{tabular}
    \caption{ Training and inference time comparison with CIFAR-10against CIFAR-100 (OOD) detection on ResNet-34. (Also see~\cref{fig:latency})}
    \label{tab:latency}
\end{table}

\begin{table*}[!t]
\centering
\resizebox{\textwidth}{!}{
\begin{tabular}{ccccccc}
\toprule
     \multirow{2}{*}{\begin{tabular}[c]{@{}c@{}} In-dist \\ (model) \end{tabular}}  &   \multirow{2}{*}{OOD}      &    Energy~({\color{green}\textbf{w/o}} FT)       & Gram~({\color{green}\textbf{w/o}} FT)  & G-ODIN~({\color{red}\textbf{w/}} FT)  &1D~({\color{red}\textbf{w/}} FT)  & Ours-MLP~({\color{green}\textbf{w/o}} FT)           \\ 
     \cline{3-7} 
                                &                                & \multicolumn{5}{c}{ TNR at TPR 95\% / AUROC  / Detection acc. }                \\ 
\midrule
\multirow{6}{*}{\begin{tabular}[c]{@{}c@{}} CIFAR-10 \\(ResNet-34) \end{tabular}}
&  iSUN
& \multicolumn{1}{c}{60.4 / 92.2 / 87.0}
& \multicolumn{1}{c}{99.3 / 99.8 / 98.1}
& \multicolumn{1}{c}{95.3 / 98.9 / 95.6}
& \multicolumn{1}{c}{76.9 / 86.3 / 92.9}
& \multicolumn{1}{c}{\textbf{99.7 / 99.9 / 98.6}} \\
&  SVHN
& \multicolumn{1}{c}{58.4 / 90.6 / 85.5}
& \multicolumn{1}{c}{97.6 / 99.5 / \textbf{96.7}}
& \multicolumn{1}{c}{89.5 / 97.8 / 92.9}
& \multicolumn{1}{c}{86.2 / 95.1 / 88.9}
& \multicolumn{1}{c}{\textbf{97.7} / \textbf{99.6} / 96.6} \\
&  Texture
& \multicolumn{1}{c}{41.1 / 85.5 / 80.8}
& \multicolumn{1}{c}{88.0 / 97.5 / 91.9}
& \multicolumn{1}{c}{81.4 / 95.0 / 88.9}
& \multicolumn{1}{c}{72.4 / 91.1 / 84.9}
& \multicolumn{1}{c}{\textbf{ 94.0 / 98.9 / 94.6}} \\
&  LSUN-C
& \multicolumn{1}{c}{89.2 / 98.0 / 93.8}
& \multicolumn{1}{c}{89.8 / 97.8 / 92.6}
& \multicolumn{1}{c}{\textbf{93.9 / 98.8} / 94.0}
& \multicolumn{1}{c}{77.1 / 92.9 / 86.5}
& \multicolumn{1}{c}{\textbf{93.9 / 98.8 / 94.5}} \\
&  ImageNet-C
& \multicolumn{1}{c}{67.4 / 93.6 / 88.7}
& \multicolumn{1}{c}{\textbf{96.7} / \textbf{99.2} / \textbf{96.1}}
& \multicolumn{1}{c}{90.8 / 98.2 / 94.3}
& \multicolumn{1}{c}{81.9 / 94.6 / 88.5}
& \multicolumn{1}{c}{96.1 / \textbf{99.2} / 95.6} \\
&  CIFAR-100       
& \multicolumn{1}{c}{43.1 / 87.1 / 80.7}
& \multicolumn{1}{c}{32.9 / 79.0 / 71.7}
& \multicolumn{1}{c}{36.3 / 85.5 / 79.3}
& \multicolumn{1}{c}{57.4 / 87.2 / 80.8}
& \multicolumn{1}{c}{\textbf{63.8 / 90.1 / 83.4}} \\

\bottomrule
\end{tabular}
}
\caption{Comparison of detection methods when only in-distribution dataset is accessible. (Also see~\cref{fig:compare_only_in})}
\label{tab:no_ood_data}
\end{table*}
\end{document}